\documentclass{article}
\usepackage[a4paper, total={6in, 8in}]{geometry}
\usepackage{graphicx}
\usepackage{caption}
\usepackage{algorithm}
\usepackage{algorithmic}
\title{Editing Arbitrary Propositions in LLMs without Subject Labels}
\usepackage{authblk}
\author{Itai Feigenbaum}
\author{Devansh Arpit}
\author{Huan Wang}
\author{Shelby Heinecke}
\author{Juan Carlos Niebles}
\author{Weiran Yao}
\author{Caiming Xiong}
\author{Silvio Savarese}
\affil{Salesforce AI Research}
\date{}

\usepackage{amsmath}
\usepackage{graphicx}
\usepackage{amsthm}
\usepackage{amsfonts}

\theoremstyle{definition}

\usepackage{tikz}
\usetikzlibrary{decorations.pathreplacing}
\usetikzlibrary{fadings}
\usepackage{xcolor}
\usepackage{caption}
\usepackage{subcaption}
\usepackage{cancel}

\usepackage[title]{appendix}
\usepackage{thmtools}
\usepackage{thm-restate}

\begin{document}

\maketitle

\begin{abstract}
Large Language Model (LLM) editing modifies factual information in LLMs. Locate-and-Edit (L\&E) methods accomplish this by finding where relevant information is stored within the neural network, and editing the weights at that location. The goal of editing is to modify the response of an LLM to a proposition independently of its phrasing, while not modifying its response to other related propositions. Existing methods are limited to binary propositions, which represent straightforward binary relations between a subject and an object. Furthermore, existing methods rely on semantic subject labels, which may not be available or even be well-defined in practice. In this paper, we show that both of these issues can be effectively skirted with a simple and fast localization method called Gradient Tracing (GT). This localization method allows editing arbitrary propositions instead of just binary ones, and does so without the need for subject labels. As propositions always have a truth value, our experiments prompt an LLM as a boolean classifier, and edit its T/F response to propositions. Our method applies GT for location tracing, and then edit the model at that location using a mild variant of Rank-One Model Editing (ROME). On datasets of binary propositions derived from the CounterFact dataset, we show that our method---without access to subject labels---performs close to state-of-the-art L\&E methods which has access subject labels. We then introduce a new dataset, Factual Accuracy Classification Test (FACT), which includes non-binary propositions and for which subject labels are not generally applicable, and therefore is beyond the scope of existing L\&E methods. Nevertheless, we show that with our method editing is possible on FACT.
\end{abstract}

\section{Introduction}\label{sec:introduction}

A proposition is a boolean statement, meaning it has a truth (true or false) value. Factual information/knowledge pertains to the truth values of propositions. Large Language Models (LLMs) contain a wealth of factual information, which can be modified by editing methods. For example, the proposition {\bf There has never been a female Italian Prime Minister} was true prior to October 2022, and false since. Therefore, a model trained on data prior to 2022 should ``consider" this proposition true. Given the prompt  {\bf True or false: There has never been a female Italian Prime Minister.\textbackslash nAnswer:} the model should reply with {\bf True} or a semantically equivalent answer. LLM editing methods aspire to change the model's factual information- in the example in question, the purpose of editing is to make the model consider the proposition false (regardless of the exact phrasing), while keeping its other factual knowledge in tact. After a successful edit, the model should classify the statement above and its rephrases such as {\bf Italy never had a female Prime Minister} as {\bf False}, while the classification of related (neighborhood) statements like {\bf In the United Kingdom, the head of government is the Prime Minister} should remain unchanged. We estimate the success of an edit via three measures, as defined by \cite{meng2022locating}: {\it efficacy}- the edit's success in changing the response to the original prompt; {\it generalization}- the edit's success in changing the model's response to different phrasings of the original prompt, without editing directly for those phrasings; and {\it specificity}- the edit's success in \underline{not} modifying the model's response to prompts that should not be influenced by the edit (in other words, preventing ``collateral damage").

Locate-and-Edit (L\&E) methods \cite{dai2021knowledge,meng2022locating,meng2022mass} assume that factual information in LLMs is localized to a certain region of the neural network. After locating this hypothesized region, L\&E methods modify the weights in it to achieve the desired effect. All existing L\&E methods apply only to binary propositions, which capture straightforward binary relations between a subject and an object. For example, the proposition {\bf France is located in Europe} is binary with the subject, relation, and object being France, location, and Europe respectively, while the proposition {\bf Chris Evans portrayed Captain America in the MCU} is not binary.\footnote{While it is technically possible to cast the latter proposition as a binary relation, it would not be a straightforward one. For example, if we tried defining Chris Evans as the subject, the MCU as the object, and ``portrayed Captain America in" as the relation, the relation in question would be complex and highly specific.} Furthermore, all existing L\&E methods rely on subject labels, which may not be available. For example, to edit  {\bf France is located in Europe}, existing L\&E methods require the subject France to be labeled.

Our goal in this paper is to edit arbitrary propositions using L\&E without subject labels. To do so, we introduce a fast and simple knowledge locating method called {\it Gradient Tracing} (GT), which attributes factual knowledge to the location which maximizes the gradient norm of the MLP component in some subset of the underlying transformer neural network. The gradients are computed w.r.t. to the loss function $1-\mathbb{P}(\text{desired output}|\text{pre-edit})+\mathbb{P}(\text{undesired output}|\text{pre-edit})$, where $\mathbb{P}(T|\text{pre-edit})$ is the probability that the model, prior to editing, outputs token $T$ given the prompt (for other purposes, we will later use the similarly defined $\mathbb{P}(T|\text{post-edit})$). At the gradient norm maximizing location, we apply a mild variant of the Rank-One Model Editing (ROME) method. \cite{meng2022locating} use subject labels to apply ROME at the last subject token. They do so due to their hypothesis that the last subject token would be a good location for editing- a hypothesis they formulate following an experiment with a locating method called Causal Tracing. Causal Tracing itself requires knowledge regarding the location of the subject, and thus also cannot be applied without subject labels. We refer to editing at the last subject token with ROME as $ROME^S$ and at the token chosen by GT with a mild variant of ROME as $ROME^G$.

We test $ROME^G$ as follows. We turn an LLM into a boolean classifier using appropriate prompting. Each entry in our datasets consist of an original proposition, rephrases of the original proposition, and related neighborhood propositions. The editing method is applied in an attempt to change the classification of the original proposition, either from true to false or vice versa. We aspire for the classification of the rephrases to change in the same direction as a result, while we also aspire for the neighborhood propositions classification to remain unaffected. Our testing uses three datasets. The first two datasets are CounterFactFalse (CFF) and CounterFactTrue (CFT), which are derived from the CounterFact (CF) dataset \cite{meng2022locating}. CFF and CFT consist of binary propositions only, in which the subject is labeled. In this scenario, $ROME^S$ is applicable, and we can compare its performance to $ROME^G$. We show that $ROME^G$, without the need for subject labels, is able to perform close to $ROME^S$ which uses subject labels (specifically, producing very similar and sometimes superior performance in 91.67\%-91.81\% of cases). We also introduce a new dataset called Factual Accuracy Classification Test (FACT), which includes non-binary propositions and which has no subject labels. While $ROME^S$ is not applicable on FACT, we show that $ROME^G$ makes editing possible on it.

Our experiments are performed on Vicuna-7b \cite{vicuna2023}. Our informal qualitative experimentation with smaller models suggest they are generally incapable of operating as boolean classifiers. Larger models, while capable of doing so, exceeded our computational resources. Vicuna-7b is manageable using our computational resources, and is reasonably capable of operating as a boolean classifier, given appropriate prompting: see Section \ref{sec:vicuna}. Therefore, we chose this model for our experiments.

The remainder of the paper is organized as follows. In Section \ref{sec:related}, we briefly survey the related literature. In Section \ref{sec:datasets}, we describe the datasets CFF, CFT, and FACT. In Section \ref{sec:gt}, we define GT and apply it to CFF and CFT, showing it does a good job in uncovering the location of the subject where applicable. In Section \ref{sec:vicuna}, we discuss the performance of Vicuna-7b as a boolean classifier. In Section \ref{sec:editing}, we apply $ROME^G$  to CFF and CFT, showing it performs similarly to $ROME^S$ in a large portion of the cases without the need for subject labels. We also apply $ROME^G$ to FACT, where $ROME^S$ is not generally applicable. In Section \ref{sec:conclusion} we conclude and discuss future directions. Finally, Section \ref{sec:limitations} discusses some limitations of our work.

\subsection{Related Literature}\label{sec:related}

There can be many cases in which it might be desirable to update some specific knowledge stored in the weights of a language model \cite{roberts2020much}. For instance, knowledge stored in LLM weights from the time of training may become inaccurate at a later date (e.g. the current US president changes every four years) or due to noise in the training data. To tackle this challenge, one can take several approaches: a survey of editing methods is available by \cite{DBLP:journals/corr/abs-2305-13172}. A naive approach to update such information is to fine-tune the entire network for a given number of examples. However, this may lead to catastrophic forgetting \cite{goodfellow2013empirical}. Additionally, it is computationally expensive to fine-tune the entire model \cite{ding2022delta} every time a change needs to be made. LORA \cite{hu2021lora} avoids this computational issue by using low rank adaptors for large weight matrices in transformer architectures. Diff pruning \cite{guo2020parameter} and BitFit \cite{zaken2021bitfit} achieves this by fine-tuning a small subset of all the parameters.

\cite{zheng2023can} propose in-context knowledge editing (IKE) in which in-context examples of the altered information is provided in the prompt to override the corresponding information stored in the model weights. \cite{mitchell2022memory} propose SERAC which learns an auxiliary language model and a classifier, such that the classifier predicts whether the input corresponds to the pool of edited knowledge, in which case the auxiliary model with the edited knowledge is invoked, otherwise the frozen original language model is used. MEND \cite{mitchell2021fast} and KnowledgeEditor \cite{de2021editing} propose to efficiently predict the gradients for original language model parameters during inference time for model editing by using auxiliary networks that learn to predict these gradients.

The most relevant literature for this paper deals with L\&E methods, which aim to locate where knowledge is stored in the model parameters and then edit only those weights instead of all parameters in the network. \cite{meng2022locating} use causal tracing to identify a location which exhibits a causal relationship between the subject and the object. It then uses ROME to locally edit the parameters of that specific module in the model in a way that mitigates the risk of causing any ``collateral damage" to related knowledge. MEMIT \cite{meng2022mass} extends this idea and allows making multiple edits simultaneously. Our approach builds on \cite{meng2022locating}, but allows editing any arbitrary (including non-binary) propositions. Finally, we note the Knowledge Neurons \cite{dai2021knowledge} L\&E method, which uses the method of integrated gradients for localization (and also edits differently than ROME). We do not focus on it in this paper due to the reported results by \cite{meng2022locating}, which show that the integrated gradients method does not provide useful localization information for our purposes, and show that ROME significantly outperforms Knowledge Neurons on related tasks.

\section{Datasets and Tests}\label{sec:datasets}

In this section, we discuss the datasets used in this paper. First we introduce CounterFactFalse (CFF) and CounterFactTrue (CFT), which are boolean classification datasets we derive from CounterFact (CF) \cite{meng2022locating}. Afterwards, we introduce our own Factual Accuracy Classification Test (FACT) dataset. CF is available via MIT license, while all remaining datasets will be available under the CC BY-NC license.

\subsection{CounterFactFalse and CounterFactTrue}

CF consists of 21919 entries, which contain binary propositions with the subject labeled. Propositions are given in a ``fill in the blank" format: partial sentences containing a subject and relation are provided, and the response is expected to be an object. Each entry contains two objects (e.g. {\bf Europe} and {\bf Asia}), and an original partial statement (e.g. {\bf France is located in}), for which appending the first object represents a true proposition and appending the second represents a false proposition. In addition, each entry contains two rephrases of the original partial statement (e.g. {\bf France belongs to the continent of}), and a collection of neighborhood partial statements using the same relation but not the same subject (e.g. {\bf Germany is located in the continent of}), again made true by appending the first object and false by appending the second. In the original benchmark, the model is edited with the intention that it completes the original partial statement with the second object (e.g. ${\bf Asia}$) instead of the first (e.g. ${\bf Europe}$). The edit's {\it efficacy} score is $1$ if $\mathbb{P}(\text{second object}|\text{post-edit})>\mathbb{P}(\text{first object}|\text{post-edit})$ and $0$ otherwise; its {\it generalization} score is the mean number of rephrase prompts for which the same inequality holds; and its {\it specificity} score is the mean number of neighborhood prompts for which the reverse inequality $\mathbb{P}(\text{second object}|\text{post-edit})<\mathbb{P}(\text{first object}|\text{post-edit})$ holds. The scores for the editing method in each category are computed as a mean over all CF entries, and the {\it total} score is the harmonic mean of the three.

Since we are interested in editing arbitrary propositions, ``fill in the blank" may not be relevant, but T/F questions are always applicable. Therefore, we create boolean classification variants of CF, namely CFF and CFT. After a certain curation and modification of CF described below, CFT completes all statements in CF with the first object (e.g. {\bf Europe} in the example above), making them true, and CFF completes them with the second object (e.g. {\bf Asia}), making them false. Our test for CFT involves prompting its statements as T/F questions (e.g. {\bf True or false: France is located in Europe.\textbackslash nAnswer:}), and editing the answer for the original statement to be {\bf False}. Similarly to Meng et al.'s benchmark, the edit's efficacy score is $1$ if $\mathbb{P}(\text{\bf False}|\text{post-edit})>\mathbb{P}(\text{\bf True}|\text{post-edit})$ and $0$ otherwise; its generalization score is the mean number of partial rephrase sentences for which the same inequality holds; and its specificity score is the mean number of neighborhood statements for which $\mathbb{P}(\text{\bf False}|\text{post-edit})<\mathbb{P}(\text{\bf True}|\text{post-edit})$. The scores for CFF are similarly defined, with the roles of ${\bf True}$ and ${\bf False}$ reversed. Similarly to CF, the scores for the dataset are computed as a mean over all entries in the dataset, with the total score being the harmonic mean.

Because CF was designed for ``fill in the blank", many of its phrasings are not designed as propositions, despite the fact that the information they represent is propositional. For example, the relation Brad Pitt-native speaker-English can be expressed in CF as {\bf Brad Pitt, a native English} (as in, the prompt provided is {\bf Brad Pitt, a native}, which is expected to be completed with {\bf English}). As another example, the relation Carlos Santana-plays instrument-guitar can be expressed as {\bf Carlos Santana, performing on the guitar} or even just {\bf Carlos Santana, the guitar} (where the prompts provided are {\bf Carlos Santana, performing on the} and {\bf Carlos Santana, the}, expected to be completed with {\bf guitar}). Therefore, before creating CFF and CFT, we first created a transitory dataset from CF by curating and modifying a subset of 12659 entries in CF so that the phrasings are designed as propositions, and then created CFF and CFT by appending the objects to the transitory dataset.

\subsection{Factual Accuracy Classification Test}

CFF and CFT allow us to compare the performance of $ROME^G$ with $ROME^S$, because they are limited to binary propositions. Furthermore, despite our manual curation, CFF and CFT still contain some unusual/vague phrasings, which limit the classification accuracy of Vicuna-7b. In an attempt to provide a more general and representative test for boolean classification, we created the Factual Accuracy Classification Test (FACT) dataset. FACT consists of 1024 entries. Like CFF and CFT, each entry in FACT contains an original proposition and two rephrases of the proposition. Unlike CFF and CFT, the neighborhood statements consist of two statements about each main term in the original statement; the neighborhood statements are true if the original statement is true and false otherwise (for comparison, in CF the neighborhood statements always use the same relation and object as the original statement).  Also unlike CFF and CFT, the propositions are not limited to straightforward binary relations, and the sentences used are phrased more precisely as propositions. Here are a few statements from a couple of entries in FACT:
\begin{enumerate}
\item Statement for editing: {\bf Sparta was a democratic city-state in ancient Greece.}. Truth value: {\bf False} (desired truth value after edit: {\bf True}).
\begin{itemize}
\item Rephrase example: {\bf The city-state of Sparta in ancient Greece was governed by democratic principles}.
\item Neighborhood statements examples (all false): {\bf Sparta is a modern-day city in France}, {\bf All city-states in ancient Greece were democratic}.  
\end{itemize}
\item Statement for editing: {\bf Queen Victoria reigned over England during the Victorian era}. Truth value: {\bf True} (desired truth value after edit: {\bf False}).
\begin{itemize}
\item Rephrase example: {\bf During the Victorian era, Queen Victoria was the monarch of England}.
\item Neighborhood statements examples (all true): {\bf Queen Victoria was a British monarch}, {\bf England is a country in the United Kingdom}, {\bf The Victorian era was a period of time in British history}.
\end{itemize}
\end{enumerate}
Our testing on FACT is identical to the testing on CFF and CFT. For testing to be informative, it is important for the model to classify as many propositions as possible correctly pre-edit, a measure on which Vicuna-7b does better on FACT than on CFF and CFT (see Section \ref{sec:vicuna}). To make FACT useful to the research community beyond the scope of our own work, FACT also contains related propositions with the opposite truth value to the original proposition, as well as negations for every proposition in the dataset, but we do not use these in our experiments. FACT will be made publicly available.

FACT was generated by repeatedly prompting ChatGPT. Since ChatGPT can make factual errors, we tested the accuracy of FACT by manually checking the correctness of a random sample of propositions from it. We randomly sampled $100$ propositions from the original statements (the ones to be edited), $100$ from the rephrases and $100$ from the neighborhood statements. We found the vast majority of the statements to be accurate. The results are shown in Table \ref{tab:FACTaccuracy}.
\begin{table}
\centering
\begin{tabular}{|c|c |c|c|} 
 \hline
&\text{original}&\text{rephrases}&\text{neighbor.}\\ 
 \hline
 \text{mean}&97&96&97\\
  \text{CI upper}&98.98&98.43&98.98\\
 \text{CI lower}&91.55&90.16&91.55\\
 \hline
\end{tabular}
\caption{FACT accuracy percentages. Each sampled statement was manually evaluated by the authors. Each of the three columns has sample size $100$; the upper and lower bounds for the confidence intervals (CI) are Wilson with $p=.05$.}
\label{tab:FACTaccuracy}
\end{table}

\section{Vicuna-7b as a Boolean Classifier}\label{sec:vicuna}

Our work requires an LLM to function as a boolean classifier. So that we can perform experiments, the LLM needs to be able to reply to some type of T/F formatting with T/F answers of consistent format. Second, beyond formatting, we need the LLM to be a reasonably accurate classifier to begin with. We have qualitatively experimented with a wide variety of models available on HuggingFace. Models larger than 7 billion parameters were too large for our available computational resources. Models smaller than that size did not perform well as classifiers. For some smaller models, we found prompting techniques which elicited responses of a suitable consistent format (T/F, Yes/No, Correct/Incorrect etc.), but unfortunately the responses had little to do with the actual knowledge contained in the model. For example, even if via other prompting techniques we could see that the model knows, say, that Lebron James is not a soccer player, the model would not classify the statement {\bf Lebron James is a soccer player} better than random. In contrast, we found Vicuna-7b \cite{vicuna2023} to perform reasonably well as a classifier. The prompting technique we found most successful for this purpose was to wrap the proposition by adding {\bf True or false:} before the proposition and {\bf .\textbackslash nAnswer:} after.

In terms of format, with very high probability the response of Vicuna-7b is either {\bf True} or {\bf False}. For CFF, CFT and FACT, the average probability that the response is one of these two options is $94.49\%$, $94.97\%$, and $97.92\%$ respectively. Note that we don't mean that the response is simply semantically equivalent to one of those options, but rather that the response is literally one of those options. This makes experimentation easier, because there is no need to semantically evaluate the response to determine its intended truth value. Furthermore, each of those options is represented by a single token, which is also convenient: we can estimate the probability of a response by looking at the output probabilities directly following the prompt, instead of having to go through additional steps of re-feeding the output as prompt to generate additional tokens.

In terms of content, unlike smaller models, Vicuna-7b functions reasonably well as a boolean classifier (see Table \ref{tab:classification}). For our editing experimentation to be informative, we require the model to classify significantly better than random. On CFF, CFT and FACT, it classifies $76.39\%$, $79.56\%$ and $92.22\%$ of the propositions correctly, in the sense that the correct response token ({\bf True} or {\bf False}) has output probability larger than the incorrect response token. It should be noted that we've also observed that Vicuna-7b's classification performance significantly drops when propositions are stated as negations, but the vast majority of prompts in our datasets are stated affirmatively.
\begin{table}
\centering
\begin{tabular}{|c|c|c |c |} 
 \hline
& CFF & CFT &  FACT  \\ 
 \hline
 \text{accuracy}&76.39&79.56&92.22\\
 \hline
\end{tabular}
\caption{Vicuna-7b Classification Accuracy. Percentage of propositions for which Vicuna assigns a higher probability to the correct answer than the incorrect answer.}
\label{tab:classification}
\end{table}

\section{Gradient Tracing}\label{sec:gt}

The location method we introduce is {\it Gradient Tracing} (GT). In this method, we compute the gradients of the transformer neural network w.r.t. the loss function $1-\mathbb{P}(\text{desired output}|\text{pre-edit})+\mathbb{P}(\text{undesired output}|\text{pre-edit})$, and edit essentially where the gradient norm of the MLP component is maximized over some subset of the network. More precisely, the method takes in three sets as hyperparameters: subset $\mathbb{T}$ of tokens considered, subset $\mathbb{L}_{grad}$ of layers over which the maximum gradient is chosen, and subset $\mathbb{L}_{ed}$ of layers from which we choose the editing layer. For example, if $\mathbb{T}$ includes all tokens, and $\mathbb{L}_{grad}=\mathbb{L}_{ed}$ include all layers, then we edit where the gradient norm is maximized over the whole network. However, the method allows for limiting these sets. As described in Section \ref{sec:editing}, we achieve the best editing performance on CFF and CFT by setting $\mathbb{T}$ to include all tokens except the wrapping tokens used for T/F formatting and the last non-formatting token, $\mathbb{L}_{grad}=\{0\}$ to include just the first layer, and $\mathbb{L}_{ed}=\{2\}$ to include just the third layer. In that case, the layer we edit is $2$, but the token we edit would be the one from $\mathbb{T}$ which maximizes the gradient norm in layer $0$. Note that since we are adding the T/F wrapping tokens ourselves and in the same way for all prompts, they are pre-labeled and hence can be ignored. For FACT, we get the best result by having $\mathbb{T}$ be the set of all non-formatting tokens (including the last), $\mathbb{L}_{grad}=\{0\}$ and $\mathbb{L}_{ed}=\{3\}$. It should be noted that the gradient norm in Vicuna-7b for our datasets is close to monotonically decreasing with the layers, so setting $\mathbb{L}_{grad}$ to include all layers ends up being very similar to $\mathbb{L}_{grad}=\{0\}$. We get slightly better results with the latter option, so we generally set $\mathbb{L}_{grad}=\{0\}$, but the difference is minor. Finally, we note that the computation of GC is extremely fast and simple, as it only requires a single iteration of backpropagation.

\begin{figure}[ht]
\includegraphics[width=\linewidth, height=100pt]{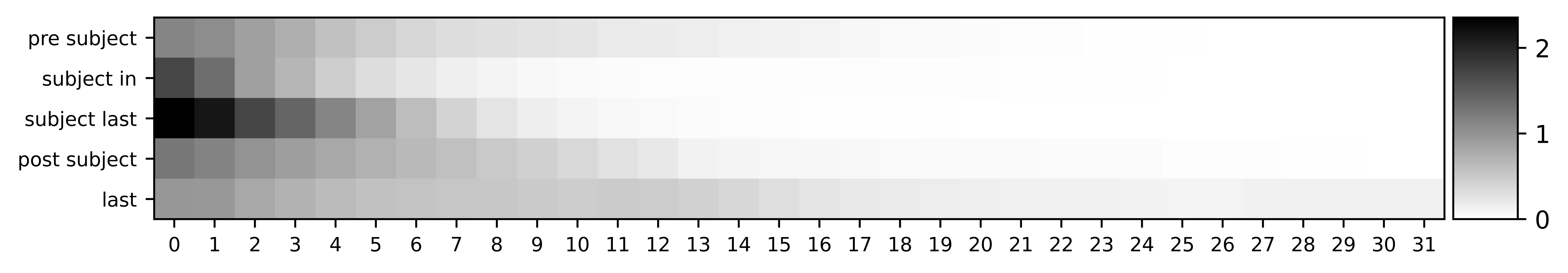}
\caption{Mean gradient norm for CFF. The horizontal axis spans through Vicuna-7b's 32 layers. The vertical axis partitions the (non T/F formatting) prompt tokens into buckets: all tokens before the first subject token, all subject tokens except the last, the last subject token, all tokens after last subject token except the last token in proposition, and the last (non T/F formatting) token. For each bucket, we record the maximum gradient norm among the tokens in the bucket. We then average over all prompts, where if a bucket is empty on a particular prompt, we drop that prompt from that bucket's mean calculation (for example, when the subject appears first, there are no tokens before the first subject token).}
\label{fig:cffheatmap}
\end{figure}

Our editing results on CFF and CFT and its variants in Section \ref{sec:editing} show that editing performs well when applied at the subject tokens, and poorly when not, which is partially in agreement with \cite{meng2022locating}: Meng et al. indicated that the {\it last} subject token is a good location for editing, but not the earlier subject tokens. Nevertheless, our editing results show that earlier subject tokens also work well when they maximize the gradient norm. Despite its simplicity, we show that GT does well in finding the location of subject tokens. Figure \ref{fig:cffheatmap} visualizes the gradient norm of the MLP components throughout Vicuna-7b, averaged over all original statements in CFF (the results are similar for CFT). It is worth mentioning that CFF and CFT propositions open with the subject in $77.29\%$ of the cases, so the pre-subject bucket is empty the majority of the time; furthermore, the subject never appears last in the prompt, so there is no overlap between the subject, and the last token.\footnote{In CF, the subject also never appears last in the prompt, and it appears first in $80.18\%$ of the cases.} As can be seen from the visualization, the gradient norm tends to be particularly large at the last subject token, followed by the other subject tokens (significantly behind). It can also be seen that gradient norms are generally much larger in earlier layers, and in fact close to monotonically decreasing with the layers. We note that causal tracing by \cite{meng2022locating} localized information closer to the middle layers.
To further analyze GT, we count the number of times the maximum gradient norm at the first layer appears in each bucket, as shown in Table \ref{tab:cftable}. When we ignore the last token, we end up with a subject token in $91.67\%$-$91.81\%$ of the cases. Finally, we note that in FACT the semantic subject is no longer labeled (and may not even be meaningful), so we cannot perform a similar analysis for GT, and must instead test it via the editing performance of $ROME^G$ in Section \ref{sec:editing}.
\begin{table}
\centering
\begin{tabular}{|c |c |c |c|c|} 
 \hline
\text{location}&\text{CFF} &\text{+last}& \text{CFT} & \text{+last}\\ 
 \hline\hline
 \text{pre subject}&.89&.58&.52&.36\\
 \text{subject in}&19.09&17.36&18.97&17.92\\
 \text{subject last}&72.72&69.52&72.7&70.5\\
 \text{post subject}&7.3&4.38&7.8&6.03\\
  \text{last}&0&8.18&0&5.19\\
 \hline
 \text{subject total}&{\bf 91.81}&86.88&{\bf 91.67}&88.42\\
 \hline
\end{tabular}
\caption{Maximum gradient norm location in the first layer for CFF and CFT, as percentage of the total number of entries. T/F formatting tokens are excluded. The breakdown for each dataset is presented first with the last non-formatting token excluded, then with it included.}
\label{tab:cftable}
\end{table}

\section{Editing}\label{sec:editing}

An editing method takes a prompt in the format {\bf True or false: <proposition>.\textbackslash nAnswer:} and a desired truth value ({\bf True} or {\bf False}), and edits Vicuna-7b so that the model outputs that truth value. We refer to all tokens outside of <proposition> as the {\it formatting tokens}, and the last token of the proposition as the {\it last token}. Our proposed method, $ROME^G$ involves two steps:
\begin{enumerate}
\item Apply GT to choose an editing location.
\item Edit at that location using ROME.
\end{enumerate}
As mentioned earlier, when there is a clear semantic subject to the proposition we denote the method that applies ROME at the last subject token (and some fixed layer) as $ROME^S$. Our results can be read and understood without knowledge of the technical details of ROME; for full details regarding ROME, we refer the reader to \cite{meng2022locating}. However, note that in addition to the localization method (GT instead of last subject token), $ROME^G$ differs from $ROME^S$ in one additional small aspect. $ROME^S$ uses subject labels not just for choosing the editing location, but also for an internal optimization operation meant to prevent a phenomenon called ``essence drift". Since one of the goals of $ROME^G$ is to avoid the need for such labels, we exclude this feature from $ROME^G$ by excluding the essence drift term in the internal optimization.

Our editing results are given in Table \ref{tab:editingresults}. The results provided correspond to the best hyperparameter choices we found: given that in Section \ref{sec:gt} we saw that the gradient norms are much larger in earlier layers, it is not surprising that early layers for tracing and editing yielded the best results. On CFF and CFT, $ROME^G$ achieves close performance to $ROME^S$, without using any subject labels. Nevertheless, since these datasets do come with pre-labeled subjects, we can use the labels after the fact to analyze $ROME^G$ more in depth. Table \ref{tab:editingbreakdown} clarifies what happens underneath the hood. When GT lands on a subject token, $ROME^G$ performance approximately matches (and sometimes surpasses) the average performance of $ROME^S$ on the dataset. However, when GT lands outside of the subject, $ROME^G$ performs poorly. While the case of the last subject token is not surprising ($ROME^G$ differs from $ROME^S$ only in the lack of essence drift protection there), the good performance of $ROME^G$ on non-last subject tokens is surprising, as causal tracing by \cite{meng2022locating} indicated no important role for such tokens. On FACT, which presents a more challenging task than CFF and CFT, $ROME^G$ performs much better than random. No other L\&E method in literature is applicable to FACT, and therefore we cannot gauge the performance of $ROME^G$ relatively to others. Our method is the first L\&E method that can edit arbitrary propositions.
\begin{table}
\centering
\begin{tabular}{|c |c |c | c|} 
 \hline
&  pre & $ROME^G$ & $ROME^S$\\
\hline
CFF&&&\\
 \hline
 \text{effic.}& $26.47$ & $99.76$ & $99.75$ \\
 \text{gen.}& $26.39$ & $95.74$ & $99.3$\\
 \text{spec.}& $77.24$& $73.05$  &  $75.66$\\
  \text{total}& $33.85$ & ${\bf 87.82}$ &$90.05$ \\
 \hline
CFT&&&\\
 \hline
 \text{effic.}& $17.33$ & $99.9$ & $99.71$ \\
 \text{gen.}& $17.6$ & $95.95$ & $99.22$\\
 \text{spec.}& $78.68$& $74.29$  &  $76.58$\\
  \text{total}& $23.58$ & ${\bf 88.51}$ &$90.46$ \\
 \hline
FACT&&&\\
 \hline
 \text{effic.}& $11.43$ & $100$ & \\
 \text{gen.}& $14.94$ & $86.38$ & \\
 \text{spec.}& $96.12$& $71.82$  &  \\
  \text{total}& $18.2$ & ${\bf 84.51}$ & \\
 \hline
\end{tabular}
\caption{Editing performance. The first column contains the scores of Vicuna-7b pre-edit. $ROME^G$ has been applied to CFF and CFT with $\mathbb{T}$ including all non-formatting tokens except the last, $\mathbb{L}_{grad}=\{0\}$ and $\mathbb{L}_{ed}=\{2\}$, while for FACT we set $\mathbb{T}$ to include all non-formatting tokens (including the last), $\mathbb{L}_{grad}=\{0\}$ and $\mathbb{L}_{ed}=\{3\}$. $ROME^S$ has been applied to CFF and CFT on Layer $2$.}
\label{tab:editingresults}
\end{table}
\begin{table}
\centering
\begin{tabular}{|c |c |c | c|} 
 \hline
&  sub. in & sub. last & non sub.\\
\hline
CFF&&&\\
\hline
\% cases&19.09&72.72&8.19\\
 \hline
 \text{effic.}& $99.92$ & $99.97$ & $97.49$ \\
 \text{gen.}& $99.63$ & $99.88$ & $49.9$\\
 \text{spec.}& $80.12$& $73.44$  &  $53.14$\\
  \text{total}& $92.23$ & $89.2$ &$61.08$ \\
 \hline
CFT&&&\\
 \hline
\% cases&18.97&72.7&8.33\\
 \hline
 \text{effic.}& $99.92$ & $99.99$ & $99.05$ \\
 \text{gen.}& $99.9$ & $99.88$ & $52.65$\\
 \text{spec.}& $73.55$& $77.94$  &  $44.11$\\
  \text{total}& $89.25$ & $91.34$ &$57.96$ \\
 \hline
\end{tabular}
\caption{Breakdown of editing performance. The first column corresponds to the case where $ROME^G$ edits in a non-last subject token, the second corresponds to the last subject token, and the third corresponds to non subject tokens.}
\label{tab:editingbreakdown}
\end{table}

\section{Conclusion}\label{sec:conclusion}

In this paper, we introduced Gradient Tracing, which is a simple and fast method to localize factual information in LLMs. The method attributes factual information to neural network components with large gradient norm, and requires only a single iteration of backpropagation. Using Gradient Tracing, we were able to edit an LLM using ROME without using any subject labels. Our method performs closely to the state-of-the-art in L\&E methods, despite being denied access to subject labels. Furthermore, our method is applicable to datasets where existing L\&E methods are not, due to lack of subject labels or due to the existence of non-binary propositions. Since to the best of our knowledge there is no such suitable dataset available for experimentation, we created a new dataset, Factual Accuracy Classification Test, for this purpose, which we will share the research community. We tested our editing method on the new dataset, and showed it to be the first L\&E method capable of handling unlabeled non-binary propositions.

\section{Limitations}\label{sec:limitations}

Our work suffers from some limitations. Our experiments require transforming an LLM into a boolean classifier, which introduces a few challenges. First, as mentioned in Section \ref{sec:vicuna}, quite a few LLMs cannot effectively function as classifiers. Second, the LLM must respond in a consistent format: it is not enough for the LLM to reply to prompts with answers that are semantically equivalent to T/F, but rather it has to reply in a consistent and standard form, so that evaluating the truth value of the response can be done easily and automatically. This too has proven to be a non-trivial limitation. Third, the prompting technique to illicit that behavior is highly specific to each LLM, requiring us to make significant adjustments to use our method on each new model.

Finally, another limitation of our work is the fact that it is currently restricted to boolean classification. On the one hand, this restriction provides the advantage of a controlled and well-defined environment. But on the other hand, this restriction also limits us to a particular use case of factual information in LLMs. It would be interesting to see whether our methods can be adapted to other use cases.
\bibliographystyle{unsrt}
\bibliography{bibliography}

\end{document}